\documentclass[10pt,twocolumn,journal]{IEEEtran}
\usepackage{graphicx,amssymb,amsmath,amsthm,cite}

\usepackage{color,multirow}

\def\be{\begin{equation}}
\def\ee{\end{equation}}
\def\ben{\begin{eqnarray}}
\def\een{\end{eqnarray}}

\def\D{\mathcal{D}}
\def\R{\mathbb{R}}

\def\V{\mathbb{V}}
\def\F{\mathrm{F}}

\def\vI{\mathbf{I}}
\def\vIt{\tilde{\mathbf{I}}}
\def\It{\tilde{I}}

\def\vd{\mathbf{D}}
\def\vc{\mathbf{c}}

\def\vB{\mathbf{B}}

\def\vRe{\mathbf{R}}

\def\vd{\mathbf{D}}
\def\vD{\vd}

\def\op{\hat{P}}

\newcommand{\la}{\langle}
\newcommand{\ra}{\rangle}

\def\Nx{N_x}
\def\Ny{N_y}

\newcommand{\Spann}{{\mbox{\rm{span}}}}

\title{Hierarchized block wise image approximation by 
 greedy pursuit strategies}
\author{Laura Rebollo-Neira, Ryszard Macio{\l} and Shabnam Bibi
\footnote{Copyright (c) 2012 IEEE. Personal use of this material is permitted. However, permission to use this material for any other purposes must be obtained from the IEEE by sending a request to pubs-permissions@ieee.org.
}\\
Mathematics Department, Aston University,
Birmingham, B4 7ET, UK}
\begin{document}
\maketitle 
\begin{abstract}
An approach for effective implementation of greedy selection methodologies, 
to approximate an image partitioned into blocks, is proposed.
The method is specially designed for approximating  
partitions on a transformed image. It evolves by 
selecting, at each iteration step, 
i) the elements for approximating each of the blocks
partitioning the image and ii) the hierarchized
sequence in which the blocks are approximated to reach
the required global condition on sparsity.
\end{abstract}
\section{Introduction}
The representation of an image by a piece of data, of
lower dimensionality than the pixel$/$intensity values, 
is called a sparse representation of a
compressible image.
Approximations using a redundant set, called a {\em{dictionary}}, 
may attain high sparsity, thereby
benefiting applications that
range from denoising to compression
\cite{MES08,Ela10,WMM10,SMF10,YM05,SE11,MYN11,BRN11,RNBC12}.
Those techniques for approximation which evolve by selection of
dictionary elements, called {\em{atoms}}, 
are refereed to as greedy strategies. 
The motivation of this Communication is to
consider an effective implementation of the
Orthogonal Matching Pursuit (OMP)
strategy \cite{PRK93} to approximate an image 
in the wavelet domain.  For this to be computationally effective, 
 with respect to  speed and  storage demands, 
the transformed image should  be partitioned 
into small blocks.

Approximations in the wavelet domain have been shown 
useful for image compression
\cite{YM05,SE11,MYN11}. Indeed, as will be illustrated here, 
the approximation of images which are compressible in the wavelet domain
results significantly sparser if carried out in such a domain.
In addition, while some other artifacts can be caused,
approximations by blocking in the wavelet domain
avoid visually unpleasant blocking artifacts in the intensity
image. 

The letter is organized as follows: 
Sec.~\ref{secNeed} discusses the need for 
considering a dedicated version of the OMP strategy 
to operate on a transformed image.
Sec.~\ref{secHBW} proposes a particular version, which is 
termed hierarchized block wise OMP.
The advantage of the proposed approach is illustrated  by 
numerical tests in Sec.~\ref{secNT}. The conclusions are summarized in 
Sec.~\ref{secCon}.

\section{The need for dedicated pursuit strategies to operate 
on partitions} 
\label{secNeed}
Approximating an image partitioned into blocks implies
having to consider some `communication' between the blocks. 
The need for communication appears when trying to decide up to 
what error each block is to be approximated.  
As the following discussion suggests, an appropriate stopping
criterion for recursive approximation of blocks in the
wavelet domain needs to be a {\em{global measure}}.

Firstly let us introduce the notational convention: 
$\R$ represents the sets of real  numbers.
 Boldface fonts are used to indicate Euclidean
vectors or matrices,
whilst standard mathematical fonts indicate the components,
e.g., $\vc \in \R^N$ is a vector of components
$c(i),\, i=1,\ldots,N$  and $\vI \in \R^{\Nx \times \Ny}$
a matrix of elements $I(i,j),\,i=1,\ldots, \Nx, \, j=1,\ldots,\Ny$.
In particular $\vI \in \R^{\Nx \times \Ny}$ will indicate an
intensity image of $\Nx \times \Ny$ pixels and
$\vIt \in \R^{\Nx \times \Ny}$ its corresponding
wavelet transform. For square matrices i.e., when  $\Nx=\Ny$, to 
shorten notation the range of indices is indicated as
$i,j=1,\ldots, \Nx.$

Consider, without loss of generality, a real, orthogonal,
and separable wavelet transform, 
which transforms an intensity image $\vI \in \R^{\Nx \times \Ny}$ into
the transformed image $\vIt \in \R^{\Nx \times \Ny}$. Thus,
each of the
points $\It(n,m),\, n=1,\ldots,\Nx,\,m=1,\ldots,\Ny$ 
is a Frobenius inner product
$\It(n,m)= \la \Psi_{u_n}\otimes \Psi_{v_m}, \vI \ra_F$, 
where the operation $\otimes$ indicates the tensor product. The 
 subscript $u_n$ in $\Psi_{u_n}$
is used to identify an ordered pair consisting of the scale and translation
parameters of the corresponding mother wavelet $\Psi$ . Accordingly,
for $n=1,\ldots,\Nx,\,m=1,\ldots,\Ny$,
\be
\label{Itp}
\It(n,m)= 
\sum_{i,j=1}^{\Nx,\Ny} 
\Psi_{u_n} (i) I(i,j) \Psi_{v_m}(j).
\ee
We shall assume for simplicity
a uniform partition and consider 
that a transformed image, $\vIt$, is the composition of $Q$ identical
and disjoint blocks $\vIt_q,\,q=1,\ldots,Q$. Hence, 
$\vIt = \cup_{q=1}^{Q} \vIt_q,$
where every $\vIt_q$ is a 2D block of size
$N_b \times N_b$.

It is clear from \eqref{Itp} that points of the transformed
image corresponding to a particular block of size 
$N_b \times N_b$ contain some  global information about the
intensity image.
This suggests that a suitable stopping criterion for
the approximation of a partition in the wavelet 
domain needs to be a global measure.
Consequently, even if the approximation of each block, as such, is 
carried out independently of the others, 
a greedy selection strategy implemented 
by approximating blocks of the transformed image should aim at
 selecting
i) the elements for approximating each of the blocks 
in the partition and ii) the hierarchized 
sequence in which the blocks should be approximated to reach 
the global condition required by the algorithm. The method 
introduced in the next section, which we term hierarchized 
block wise OMP (HBW-OMP) implements the selection of 
 i) and ii) simultaneously.
\section{Hierarchized Block Wise OMP}
\label{secHBW}
Given a redundant 2D dictionary $\D$ of $M$ atoms, 
$\D =\{\vd_n \in \R^{N_b \times N_b}\}_{n=1}^{M}$, 
each of the $Q$ blocks $\vIt_q \in \R^{N_b \times N_b}$ 
in the partition of the transformed image 
is approximated by an atomic decomposition 
$\It^{k_q,q}(i,j),\, i,j=1,\ldots,N_b$  
of the form
$$
\It^{k_q,q}(i,j)=\sum_{n=1}^{k_q}
c^{k_q,q}(n)D_{\ell^q_n}(i,j),
\, q=1,\ldots,Q.$$
For each $q$, the atoms 
$\vd_{\ell^q_{n}},\,n=1,\ldots,k_q$  
are selected from the dictionary  $\D$ 
as follows:   
On setting $\vRe^{0,q}=\vIt_q,\, q=1,\ldots,Q$
at each iteration the algorithm selects the atom     
$\vd_{\ell^{q}_{k_q+1}}$
that maximizes the absolute value of the Frobenius inner products
$\la \vd_{n} ,\vRe^{k_q,q}\ra_\F,\, n=1,\ldots,M,\,q=1,\ldots,Q$, i.e.,
\be
\begin{split}
\label{omp}
\ell_{k_q+1}^{q}&=
\operatorname*{arg\,max}_{\substack{n=1,\ldots,M\\
q=1,\ldots,Q}} | \la \vd_{n} ,\vRe^{k_q,q}
 \ra_F|\\
\text{with}\\
\vRe^{k_q,q}& = \vIt_q- \sum_{n=1}^{k_q} c^{k_q,q} (n) 
\vd_{\ell_n^{,q}}.
\end{split}
\ee   
For each $q$ and $k_q$ the coefficients $c^{k_q,q}(n),\,n=1,\ldots,k_q$
in \eqref{omp}
are such that $\|\vRe^{k_q,q}\|_\F$ is minimum, where $\|\cdot \|_\F$ is
the norm  defined  through the Frobenius inner product.
This is ensured by
requesting that $\vRe^{k_q,q}= \vIt_q- \op_{\V_{k_q}^q} \vIt_q$,
where $\op_{\V_{k_q}^q}$ is the orthogonal
projection operator
onto $\V_{k_q}^q=\Spann\{\vd_{\ell^{q}_n}\}_{n=1}^{k_q}$. 
The implementation discussed
in \cite{RNL02}   
provides us with the
representation of $\hat{P}_{\V_{k}^q} \vIt_q$ as given by,
$$\hat{P}_{\V_{k}^q} \vIt_q = \sum_{n=1}^{k_q} 
\vD_{\ell^{q}_n} \la \vB_n^{k,q}, \vIt_q \ra_F =
\sum_{n=1}^{k_q}  c^{k,q}(n)  \vd_{\ell^{q}_n}.$$
For $q$ fixed, the matrices  $\vB_{n}^{k_q},\,n=1,\ldots,k_q$ 
are biorthogonal to the selected atoms $\vd_{\ell^{q}_n}, \,n=1,\ldots,k_q$
and span the identical subspace $\V_{k_q}^q$. 
These matrices can be effectively calculated by recursive 
biorthogonalization and Gram Schmidt
orthogonalization with one re-orthogonalization step.
The coefficients in \eqref{omp} are obtained from the
inner products
$c^{k_q,q}(n)= \la \vB_{n}^{k_q,q}, \vIt_q \ra_\F,\, n=1,\ldots,k_q.$ 

For a given number $K$, the
algorithm iterates until the condition $\sum_{q=1}^Q k_q=K$,
is met. In other words, 
 the algorithm stops when the maximum number of {\em{total  atoms}}  
 required for the image approximation is reached. 

The difference between OMP and the hierarchized  block wise 
(HBW) version discussed here lies in a) the sequence in which the 
blocks are {\em{partially}} approximated  at each step  and
b) the stopping criterion. 
Standard OMP would be applied independently to each block 
 up to a given error, which is independent of the approximation 
of the other blocks. 
As condition \eqref{omp} states,
HBW-OMP adds a hierarchized selection of the
blocks to be approximated in each iteration. 
Thus, at each step, the maximum in  \eqref{omp} changes only 
for the selected block.
This implies that, by storing the maximum for each block,
the  selection of blocks introduces 
the overhead of finding the maximum element of an array of size $Q$.
As far as storage is concerned, HBW-OMP has to  
 store all the stepwise outputs of the algorithm, 
which includes matrices 
$\vB_{n}^{k_q},\,n=1,\ldots,k_q$,
for each of the $Q$ blocks in the partition. While the 
approximation is carried out in a block wise manner, the relation 
 introduced by the condition  $\sum_{q=1}^Q k_q=K$ inhibits  
 the complete approximation of each block at once.
Other pursuit strategies can also be adapted to this selection process. 
 In particular, the HBW implementation of the Matching Pursuit 
 (MP) method \cite{MZ93} differs from that of OMP in that, for each
 $q$ and $k_q$, the coefficients $c^{k_q,q}(n),\,n=1,\ldots,k_q$
in \eqref{omp} are calculated simply as 
$c^{k_q,q}(n)= \la \vd_{\ell^{q}_n}  ,\vRe^{k_q-1,q}\ra_F$. 

It is appropriate to highlight also the difference between the
proposed approach and the  Block OMP$/$MP (BOMP$/$BMP)
approach introduced in \cite{EKB10}. While BOMP$/$BMP
selects {\em{blocks of atoms}} at each iteration step,
our proposal keeps selecting the atoms as in OMP$/$MP , 
i.e., one by one.
\section{Numerical tests}
\label{secNT}
The dictionary used for the tests is a 
separable mixed dictionary consisting of two components,  
$\D_1$ and $\D_2$. The component
$\D_1$ is a Redundant Discrete Cosine (RDC) dictionary
given by:
$$\D_1=\{w_i\cos(\frac{\pi(2j-1)(i-1)}{2M}),\,j=1,\ldots,N\},$$
with $w_i,\,i=1,\ldots,M$ normalization factors.  The number $N$
is set equal to the number of pixels in one of the  block sides, and
$M=2N$ to have redundancy two.  
$\D_2$ is the standard Euclidean basis, also called the Dirac basis (DB), 
i.e.
$$\D_2=\{e_i(j)=\delta_{i,j},\,j=1,\ldots,N\}_{i=1}^{N}.$$
The required 2D dictionary is  the tensor product 
$\D \otimes \D$, where $\D= \D_1 \cup  \D_2$. 
This small  dictionary,  
called RDCDB in \cite{BRN11}, is chosen only for simplicity. 
Extending the RDCDB dictionary by adding wavelet atoms, for instance 
\cite{RNB13}, or using other dictionaries such as those arising by
learning processes \cite{YBD09,SE10,TF11}, would result 
in an increment of sparsity with all the approaches. However
the purpose of these numerical tests is only to illustrate the advantage 
of implementing matching pursuit strategies on partitions, 
in the proposed HBW manner.
\begin{figure}
\begin{center}
\hspace{-0.5cm} 
\includegraphics[width=2.5cm]{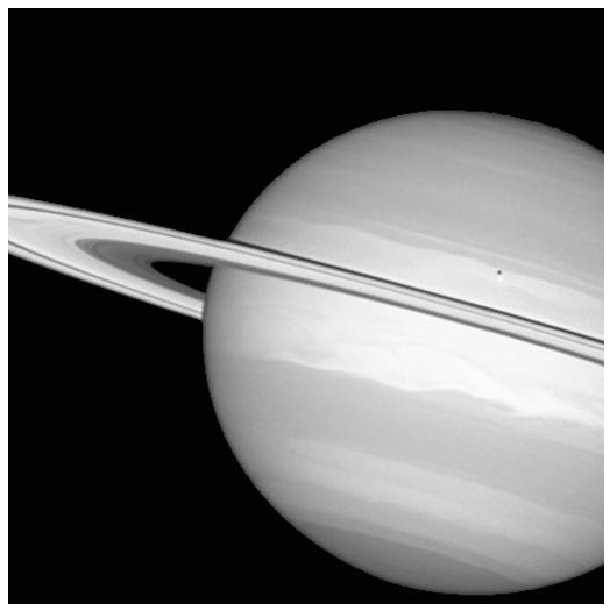}
\includegraphics[width=2.5cm]{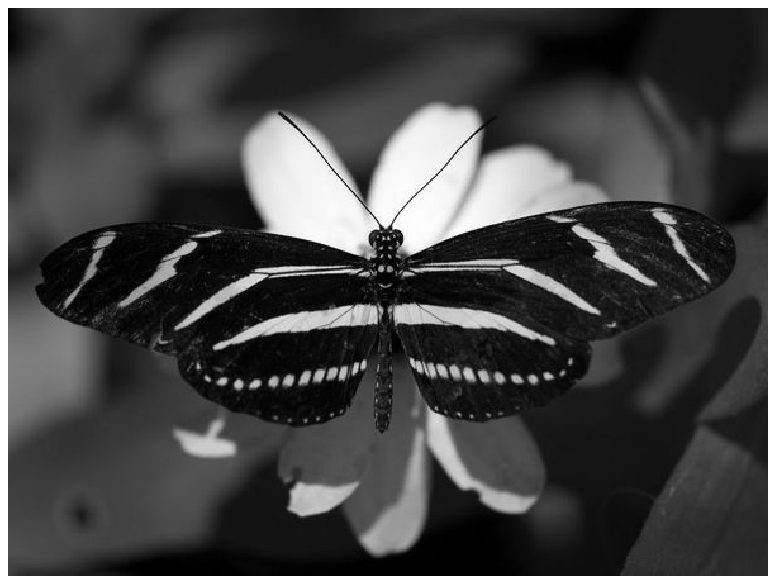}
\includegraphics[width=2.5cm]{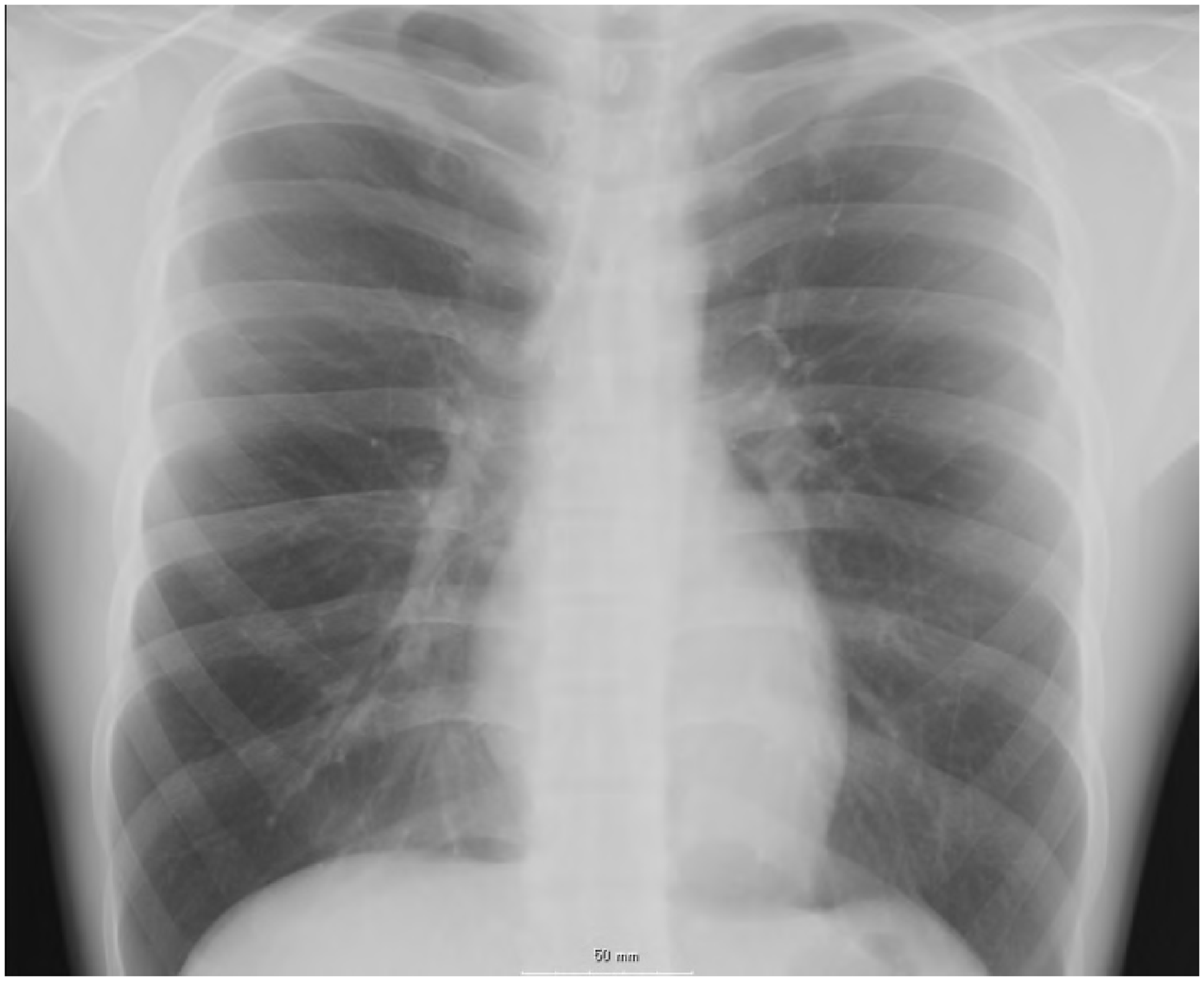}\\
\hspace{-0.1cm}
\includegraphics[width=2.3cm]{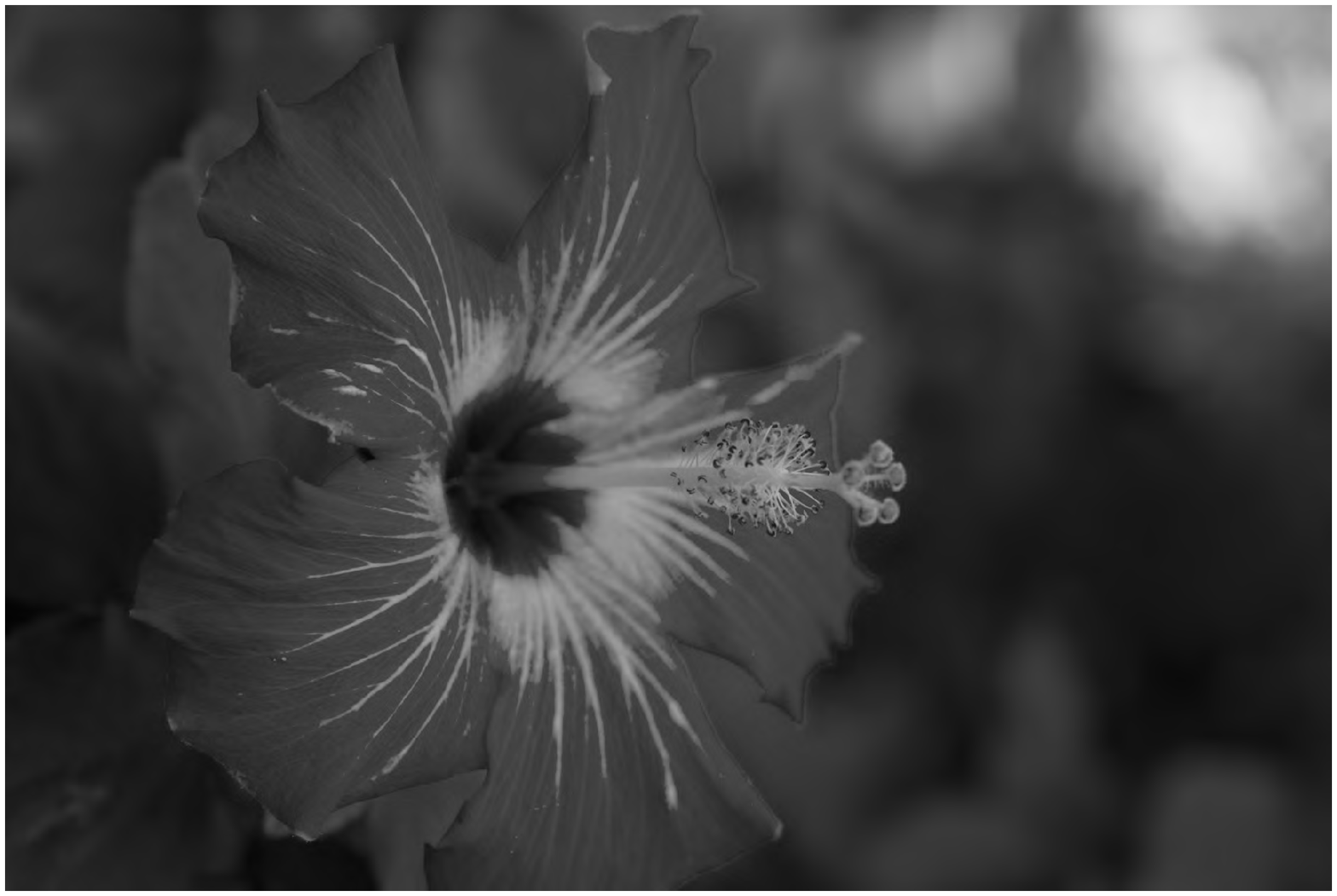}
\includegraphics[width=2.3cm]{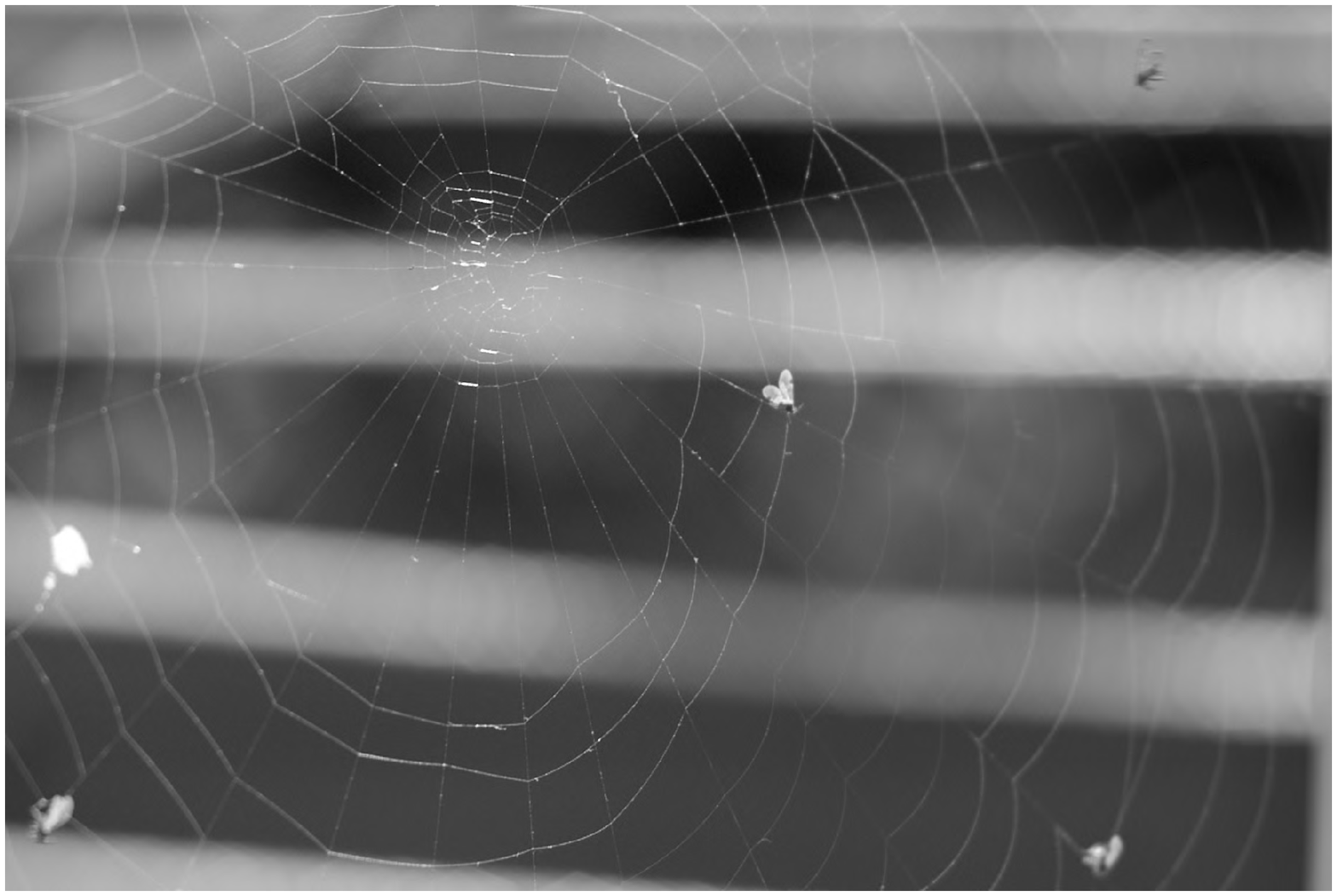}
\includegraphics[width=2.3cm]{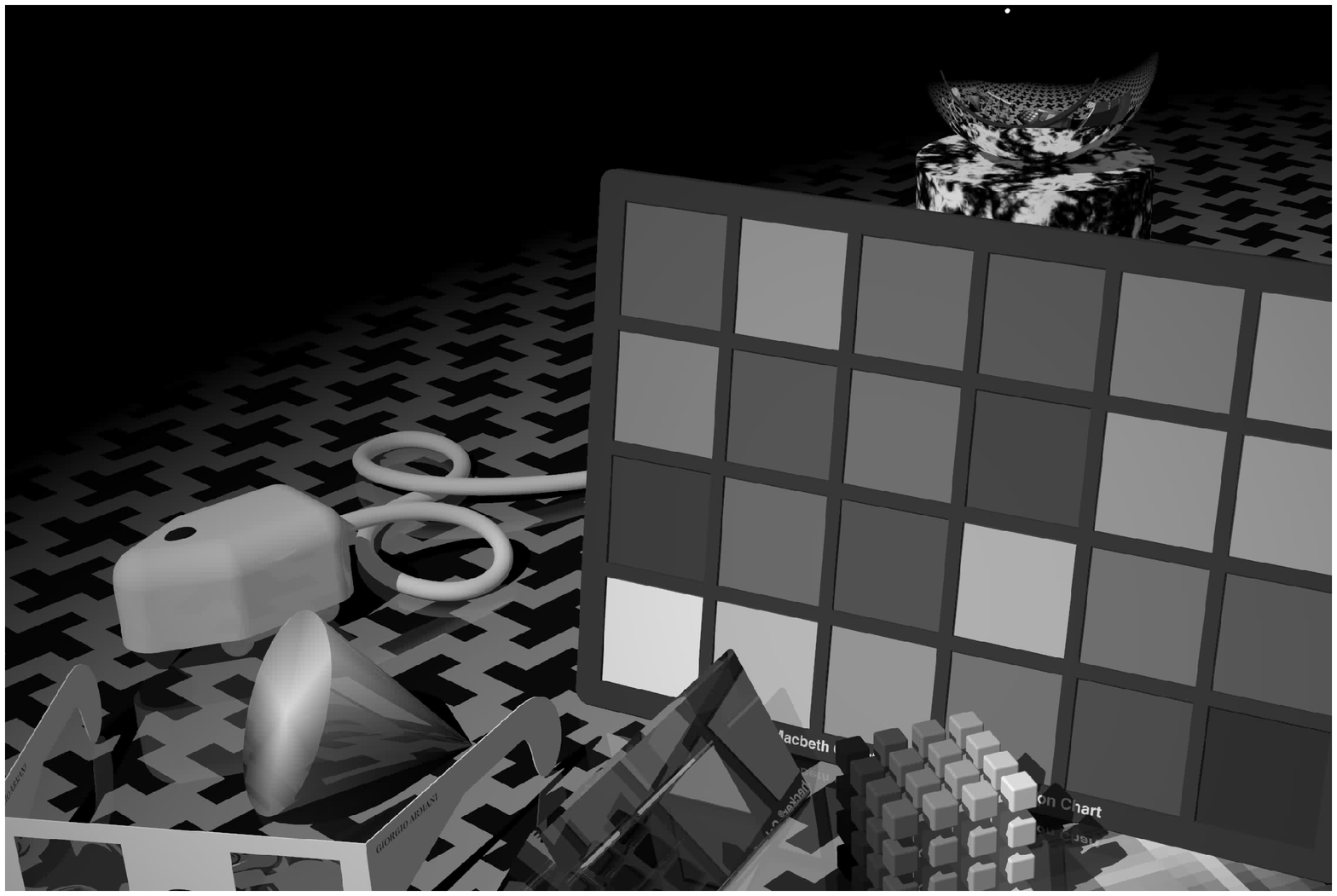}
\end{center}
\vspace{-0.5cm}
\caption{\small{First row: Planet, Butterfly and  Chest X-Ray
images of
sizes $512 \times 512$,  $448\times 600,$ and
$592 \times 728$ pixels, respectively. 
Second row: Flower, Spider Web and  Artificial high resolution
images from \cite{webcompression}, all of size $1512 \times 2264$ pixels.}}
\label{test_images_id}
\end{figure}
The set of six grey intensity levels images used for 
illustrating the proposed HBW strategy are shown in Fig 1. 
The images in the first row: Planet, Butterfly and Chest X-Ray 
are of  sizes 
$512 \times 512$,  $448\times 600$ and $592 \times 728$ pixels. 
The images in the second 
row: Flower, Spider Web and Artificial are  high resolution images 
taken from \cite{webcompression}. The Spider Web and Artificial are 
both only fractions of the much larger images in 
\cite{webcompression}, which 
we have cropped to the same size as the  
Flower, i.e., $1512 \times 2264$ pixels.

All the images  of  Fig. 1 are
approximated in both, the intensity and wavelet domains by 
partitions of $8 \times 8$ pixels. 
The transformed image is obtained using the 
CDF97 WT. 
The sparsity of an approximation is measured by the 
Sparsity Ratio (SR) defined  as 
$\text{SR}= \frac{N_x N_y}{K}$, 
with $N_x N_y$ the total number of 
pixels and  $K$  the number of nonzero coefficients to 
approximate the whole image.
\begin{table}[!h]
\label{tab1}
\begin{center}
\begin{tabular}{|c|c|c|c|c|c|c|}
\hline
I&MP&HBW&OMP&HBW&WT&DCT\\ \hline \hline
P&28.1&29.0&30.4& 31.0 &40.7&27.1 \\ \hline
B&10.6&11.2&12.3& 12.7 &9.6&10.7  \\ \hline
C&22.6&23.2&23.6& 24.4  & 34.2& 20.7  \\ \hline
F&41.0&42.8&42.6& 43.9  & 119.5&40.0 \\ \hline
S&34.8& 35.7  &36.5& 37.0& 70.0& 33.5  \\ \hline
A&26.4 & 27.5 &29.7& 30.7& 24.9& 19.4 \\ \hline \hline
\end{tabular}
\caption{{SR resulting from approximating (in the intensity domain)
the six images of Fig. 1,
up to PSNR=45.0dB, by the methods:  MP, OMP, their corresponding
HBW versions,  WT and DCT.}}
\end{center}
\end{table}
\begin{table}[!h]
\label{tab2}
\begin{center}
\begin{tabular}{|c|c|c|c|c|c|c|}
\hline
I&MP&HBW&OMP&HBW& WT&DCT\\\hline \hline
P&39.1&53.0&43.9& 62.9 &40.7& 27.1  \\ \hline
B&10.6& 12.0&12.7&14.4&9.6& 10.7\\ \hline
C&32.5&53.2&35.0& 60.7&34.2& 20.7  \\ \hline
F&48.6&156.6&50.5& 181.5&119.5& 40.0\\ \hline
S&41.1&87.9&43.3& 99.2 &70.0& 33.5\\ \hline
A&25.0&30.1&28.4& 35.0&24.9& 19.4 \\ \hline \hline
\end{tabular}
\caption{Same description as in Table 1 but
in this case MP, OMP and the corresponding HBW versions
are implemented in the wavelet domain. The results of the  WT and DCT 
are repeated here to facilitate the comparison.}
\end{center}
\end{table}
The approximation of all the images is carried out to achieve the 
required PSNR of 45.0dB using the above 
defined RDCDB dictionary and the greedy strategies
OMP, MP and their corresponding HBW versions. The 
SRs arising when these techniques are implemented in the 
intensity domain are displayed in the first 
four columns of Table 1. The rows, from top to ground, 
correspond to the Planet (P), Butterfly (B), Chest (C), 
Flower (F), Spider Web (S) and Artificial (A) images (I). 
For comparison purposes, the results produced by  
conventional WT and DCT approaches are displayed 
in the last two columns of the table. The WT results
are obtained by applying the CDF97 WT, on the whole image, and 
reducing coefficients by iterative thresholding 
until the PSNR of 45.0dB is reached. The results from the 
orthogonal DCT are obtained, from a partition  
of $8 \times 8$ pixels, 
also by thresholding of coefficients.

From Table 1 it is clear that: a) Except for the 
Butterfly and Artificial images, in the intensity domain 
the results produced by a
conventional WT approach significantly over perform the 
other approaches.
b) Whilst the  block independent OMP and MP methods produce smaller 
SRs than their corresponding HBW versions, the difference is not 
very significant. However, as can be observed in Table 2,  
all this is 
reversed if the greedy techniques are implemented with the 
RDCDB  dictionary but in the wavelet domain. 
In this domain: a) for all the images the HBW versions of 
MP and OMP,  
with the simple RDCDB dictionary, significantly over-perform
conventional WT and DCT approaches.
b) Except for the Butterfly and Artificial images the difference in 
the SR produced by the HBW approaches, with respect to the 
block independent versions, is {\underline{massive}}.

Putting aside the Butterfly and Artificial images, 
 the common feature of the other images in  Fig. 1 is 
the very significant difference in the SR  produced  
by a conventional WT approximation,
with respect to the conventional DCT. This is an indication that the 
representation of the images is  particularity  sparse
in the wavelet domain.
It is clear from Table 2 that, the sparser the approximation by 
the WT is, the lower the performance of OMP with respect to 
the proposed HBW version. 

Finally, some information with regard to processing 
times is relevant.
The tests presented here were run on a laptop 
with a 2.2 GHz Intel Processor and 3GB of RAM. 
The processing time (average of 10 independent 
runs in Matlab) for approximating the first three images of 
Fig 1 in the wavelet domain are: For the Planet
2.04 secs  with  OMP and 
2.78 secs with HBW-OMP. For the Butterfly 
7.13  and 11.56 secs  and  for the Chest
4.3  and 5.1 secs, respectively.
Using a C++ MEX file 
for both approaches the corresponding times are reduced up to 
ten times. Thus, the larger images 
(second row of Fig 1) were approximated using MEX files.
The running times corresponding to those images (left to right) are
5.71, 5.93 and 6.3 secs with OMP and 6.76, 9.7 and 
21.9 secs with HBW-OMP. 

In order to reduce processing time, the
approximation of a large image can be realized
by dividing  the image into segments to be processed independently.
Since each of the segments will contain different information,
the sparsity of the segments may differ from one another.
For comparison with the approximation of the whole image at once,
it is necessary to make uniform the sparsity of all the  segments.
A possible way of achieving this is to randomize  
the position of the blocks in the whole image, to ensure that 
each segment contains blocks from different regions of the image.
The implementation is carried out as follows:
i) Perform a random permutation of the small blocks in the
transformed domain.
ii) Divide the resulting image into segments.
iii) Apply HBW-OMP to each segment. iv) Place
the blocks back in the original position.

Through this simple procedure and taking 12 segment of 
   $540 \times 560$  pixels  each, the time to approximate 
the Artificial image is reduced to 6.1 secs while the  PSNR
 does not change significantly (45.0dB).
\section{Conclusions}
\label{secCon}
An approach, for HBW implementation of greedy 
methodologies operating on partitions, has been proposed. 
The proposal was motivated by 
the convenience of approximating some images in the 
wavelet domain. Certainly, if the image is characterized by having large 
smooth regions, for instance, when approximated in the 
intensity domain blocking artifacts are likely 
to be noticeable, even at high PSNR. Additionally, 
if an image is compressible in the wavelet domain, 
it can be expected to have a sparser approximation in that domain.
This was illustrated through the OMP and MP methodologies.
The numerical examples were chosen to enhance 
the fact that, for images which are highly compressible in the 
wavelet domain, 
approximations by partitions in such a  domain  may
achieve much higher sparsity,
provided that the proposed HBW version 
of greedy pursuit strategies is applied. 

\subsection*{Acknowledgements}
We are sincerely grateful to the Referees for their 
constructive criticism which has been of much
help to present the proposal in the present form.

Support from EPSRC UK is acknowledged.
The MATLAB and C++ MEX files for implementation of HBW-OMP with 
a separable dictionary (HBW-OMP2D) are available on 
\cite{webpage} (section Examples). 
\bibliographystyle{IEEEbib}
\bibliography{revbib}
\end{document}